\documentclass[12pt]{article}  
\pdfoutput=1
\usepackage[]{easymcm}  

\usepackage{pdfpages}
\usepackage{longtable}
\usepackage{tabu}
\usepackage{threeparttable}
\usepackage{listings}
\usepackage{paralist}

\graphicspath{{img/}}          

\title{Momentum Capture and Prediction System 
\\Based on Wimbledon Open 2023 Tournament Data}  

\begin{document}
\vspace{5pt}

\begin{abstract}
\begin{center}  
    Chang Liu\textsuperscript{*1}, Tongyuan Yang\textsuperscript{*1}, Yan Zhao\textsuperscript{2} \\  
    \vspace{1em}  
    \textit{School of Artificial Intelligence,Beijing Normal University} \\  
   
\end{center}

\noindent\textbf{Abstract: } There is a hidden energy in tennis, which cannot be seen or touched. It is the force that controls the flow of the game and is present in all types of matches. This mysterious force is Momentum. To capture the flow of matches and assess real-time players' performances, we propose an evaluation model combining \textbf{Entropy Weight Method(EWM)} and \textbf{Gray Relation Analysis(GRA)}. 
Then We use Mann-Whitney U test and Kolmogorov-Smirnov test, obtaining p-values of 0.0043 and 0.00128. Provide a double validation of the conclusion that the relationship between momentum fluctuations and match swings isn’t random, which suggests \textbf{momentum} has an important role in tennis match.

We develop a prediction model combining \textbf{XGBoost} and \textbf{SHAP}. We divided the raw data into training and test sets. Based on \textbf{XGBoost}, we enable the model to predict swings in games with an accuracy of 0.999013 (for multiple matches), and 0.992738 (for final). Based on \textbf{SHAP}, our model can indicate the trend and contribution of factors' effects on the prediction results, where the factor with the largest \textbf{SHAP} value is the bilateral distance ran during point, which is 0.1077 and 0.1042. 
We collect data from the four grand slams as new test sets to \textbf{test generalisability}. Although there is a slight decrease in accuracy, which are 0.9651, 0.8620, 0.9112, and 0.9467, respectively, our prediction model still has a solid ability to generalize to different scenarios of tennis matches. We consider introducing factors such as court surface and gender to improve it.
Our model is tested to be robust.Using our prediction model, we can analyze opponent's momentum swings and provide response advice for players facing new opponents.


\end{abstract}
\maketitle  
\vspace{100em}  

\section{Introduction}

\subsection{Background}
\noindent At the 2023 Wimbledon Men's Singles Final, Spanish star Carlos Alcaraz surprised fans with a stunning performance against world-class player Novak Djokovic, sparking in-depth discussions about the dynamics of the game and the players' performances.

\noindent In recent years, the sports world has become interested in the phenomenon of "momentum" in sports, which involves changes in energy and emotions during a match that can have a significant impact on the outcome of a match. But momentum is a subjective feeling that is difficult to quantify and define. The phenomenon of momentum can influence the decision-making of players and coaches, and they also have different views on momentum, some think it is random and others think it makes sense.

\noindent In order to better understand the role of momentum in the competition, researchers have tried to use data analysis and modeling techniques to study the origin, change, and impact of momentum. By analyzing match data, it is possible to identify factors that influence momentum changes and guide improved player performance and game strategies. Not only does this help with unexpected changes in the game, but it also guides the development of more effective training and competition strategies to improve the level of competition.

\begin{figure}[H]
	\centering
	\subfigure{
	\includegraphics[scale=0.30]{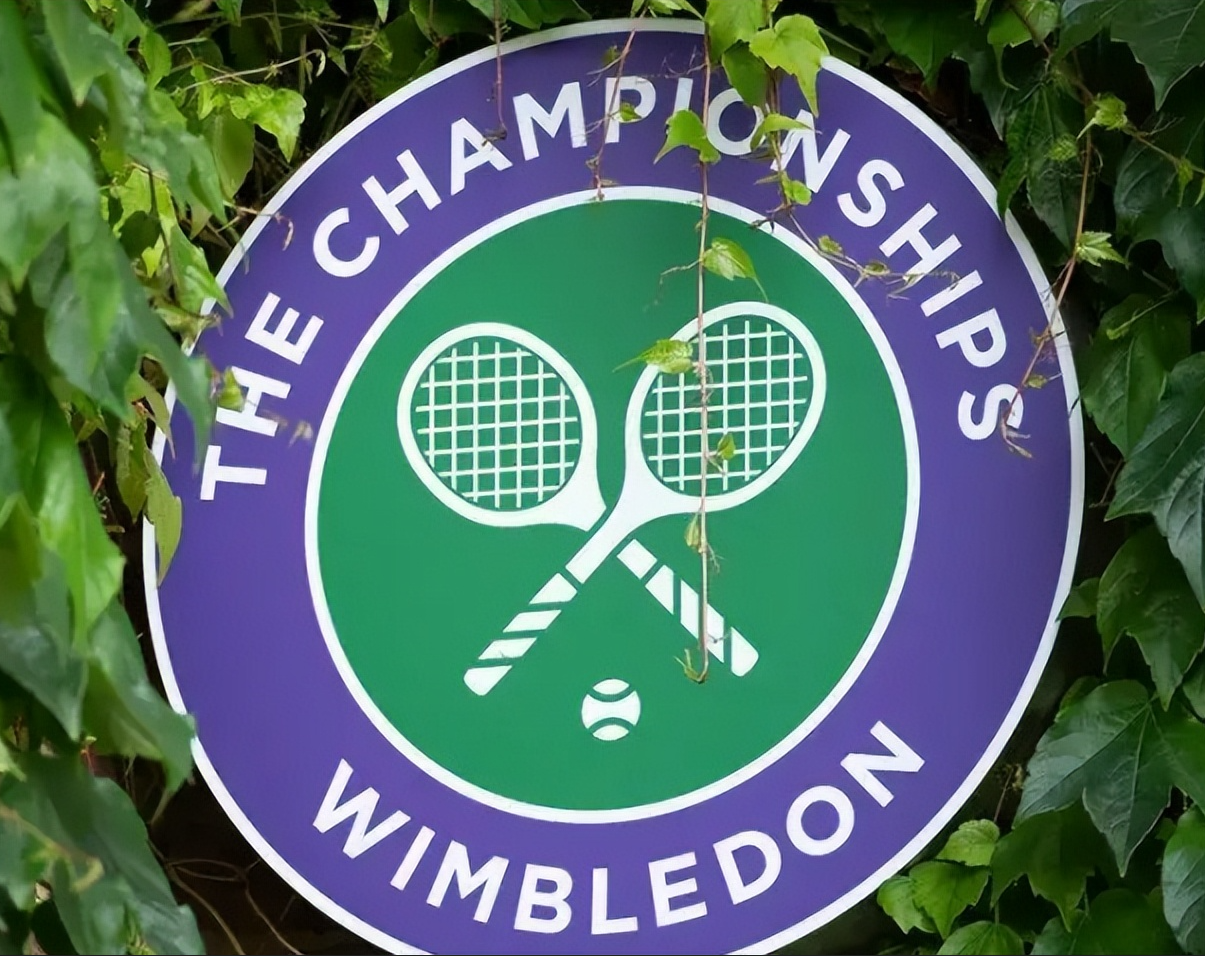}}
	\hspace{0.1in}
	\subfigure{
	\includegraphics[scale=0.25]{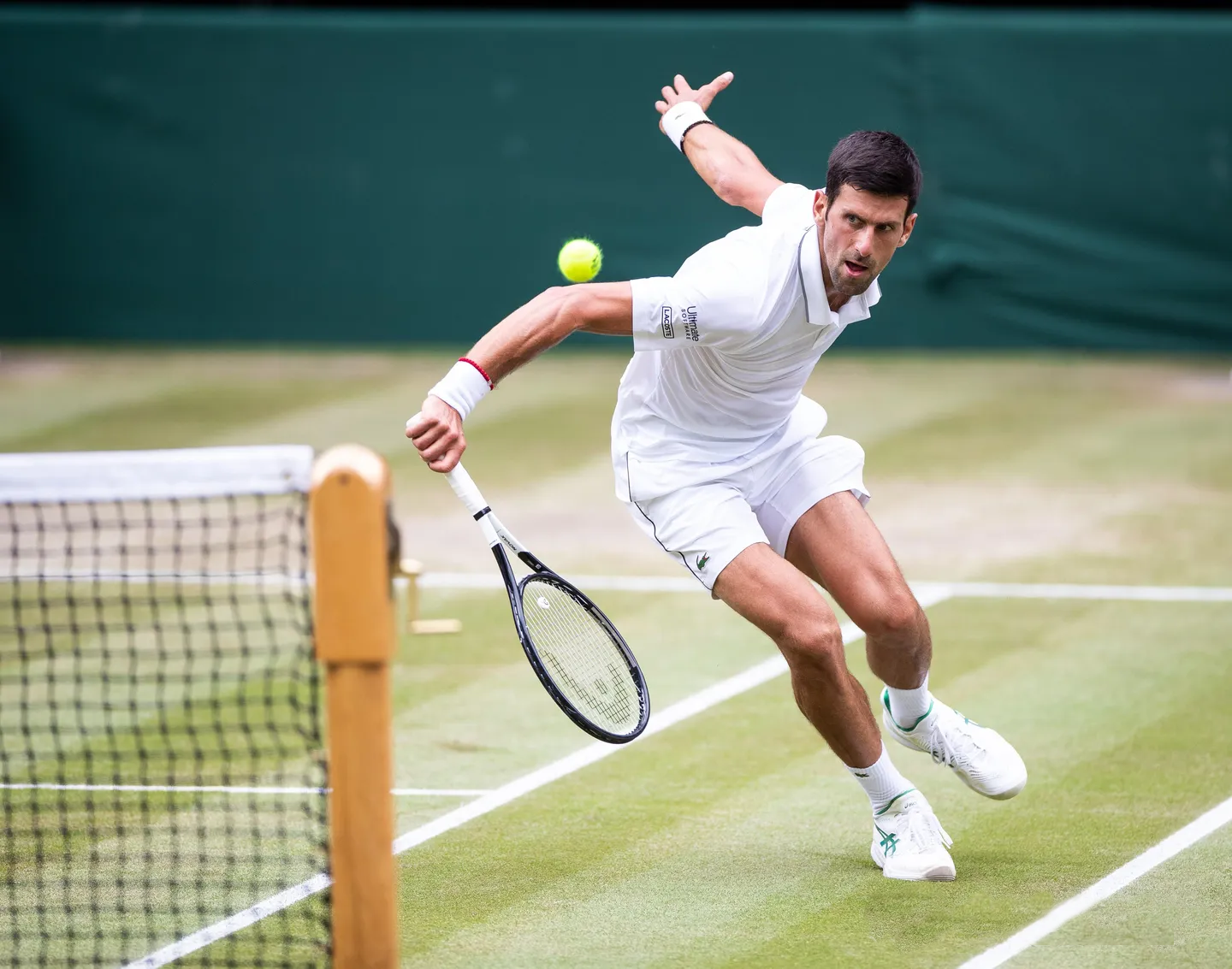}}
	\caption{Wimbledon Logo \& Pictures during Match}
\end{figure}

\subsection{Concept definition }
\noindent Momentum is the force that controls the flow of the match\textsuperscript{\cite{1}}. lt is a hiddenforce. lt is invisible because it comes from the flow of energy betweencompetitors. You can sense it when competing or spectating. lt dictates therun ofplay- you can feel things going for or against you or the players youare watching.

\noindent Momentum exists in all sports andis what makes them so exciting toplay and watch. It is why the scoredoes not always reflect the state ofplay and why the better/stronger player does not always win. Momentum gives sportunpredictability, which is whyspectators stay interested.
\noindent Here is the main idea diagram:
\begin{figure}[H]
	\centering
	\includegraphics[width=0.810\textwidth]{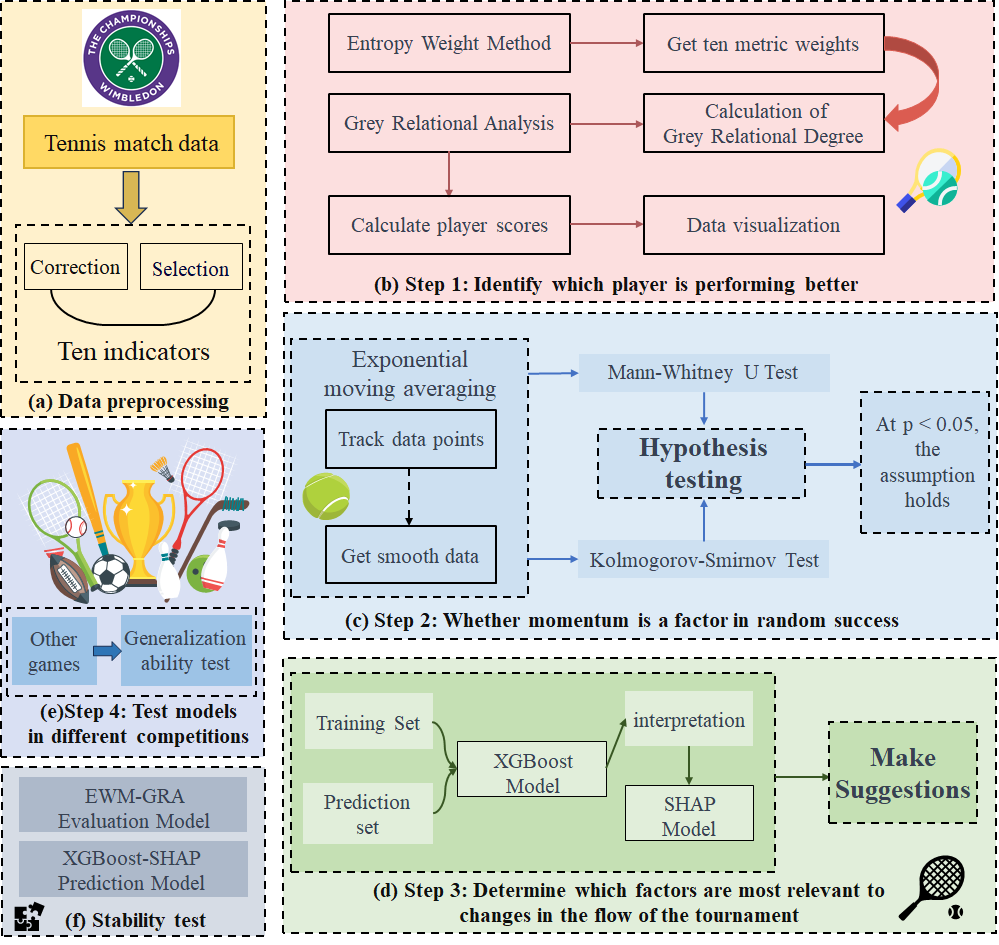}
	\caption{Workflow Chart}
\end{figure}
\section{Assumptions and Notations}
\noindent We make the following basic assumptions,each of which is adequately justified.
\begin{enumerate}[\bfseries \textit{Assumption} 1:]
	\item{\textbf{Considering that the chances of a server scoring or winning in the match is relatively higher, we introduce a serve advantage factor $\xi$ to eliminate the effect,}} when analysing the contribution of individual factors that may have an impact on a play's performance.
	\item {\textbf{The impact of off-field factors such as spectators, weather and referees are not taken into account when evaluating player performance.} In order to carry out a unified evaluation standard for all matches, they are not included in the consideration scope, which is conducive to more objective evaluation demands.}\\
	\item{ \textbf{Other historical data of players except this match are not included in the evaluation model} and only the actual trend of the game is used for evaluation and prediction.}\\
	\item {\textbf{Without considering the influence of medical suspension and rest duration on the evaluation and prediction model,} the evaluation scheme is explored only from the objective data and player performance.This assumption can lead to more accurate data analysis. }
 \end{enumerate}




\section{Model Preparation}

\subsection{Data Collection:}

\noindent With reference to IBM's relevant indicators, the following ten indicators after data Pre-processing are obtained as evaluation indicators.We then calculated these relevant metrics($x_{1}$,$x_{2}$...$x_{10}$) for all matches in the original file "Wimbledon featured matches.csv".More match statistics were obtained from the official website of Wimbledon. Detailed data sources are listed in the Table \ref{data}.

\begin{table}[H]
\begin{center}
\caption{Data sources}
\label{data}
\resizebox{\textwidth}{!}
{\begin{tabular}{c l}
\toprule[2pt]
\multicolumn{1}{m{5cm}}{\centering \textbf{Data}}
&\multicolumn{1}{m{10cm}}{\centering \textbf{Source Website} }\\ 
\midrule
Wimbledon Match Stats& https://www.wimbledon.com/en\_GB/scores/index.html \\
French Open Match Stats& https://www.rolandgarros.com/en-us/matches \\
US Open Match Stats& https://www.usopen.org/en\_US/scores/index.html \\
Australian Open Match Stats& https://ausopen.com/event-stats \\
\bottomrule[2pt]
\end{tabular}}
\end{center}
\end{table}

\noindent We first performed data cleansing on the data, and without containing missing values, we normalized the required data for subsequent calculations

\subsection{Notations}

\begin{table}[H]
\begin{center}
\caption{Notations used in this paper}
\begin{tabular}{c c}
\toprule[2pt]
\multicolumn{1}{m{3cm}}{\centering Symbol}
&\multicolumn{1}{m{8cm}}{\centering Description }\\
\midrule
$X_{1}$& Serve Advantage\\
$X_{2}$&ACE Incidence \\  
$X_{3}$ & Unforced Errors \\  
$X_{4}$ & Scoring Advantage Winning Points \\  
$X_{5}$ & Running Distance \\  
$X_{6}$& Winning Dishes and Sets \\  
$X_{7}$&  Return Depth\\ 
$X_{8}$& Serve Depth \\  
$X_{9}$& Receiving Speed\\  
$X_{10}$& Forehand Incidence \\ 
$\xi$& Serve Advantage Factor \\
$S$& Performance Score \\
$F_{1}$& P1\_Distance\_Run\\
$F_{2}$& Game\_no\\
$F_{3}$& P2\_Distance\_Run\\
$F_{4}$& P1\_score\\
$F_{5}$& Point\_No\\
$F_{6}$& P1\_Points\_Won\\
$F_{7}$& P2\_Points\_Won\\
$F_{8}$& Server\\

\bottomrule[2pt]
\end{tabular}\label{tb:notation}
 \begin{tablenotes}
        \footnotesize
        \item[*] *There are some variables that are not listed here and will be discussed in detail in each section. 
      \end{tablenotes}
\end{center}
\end{table}

\section{EWM-GRA Evaluation Model}

\begin{figure}[H]
	\centering
	\includegraphics[width=1\textwidth]{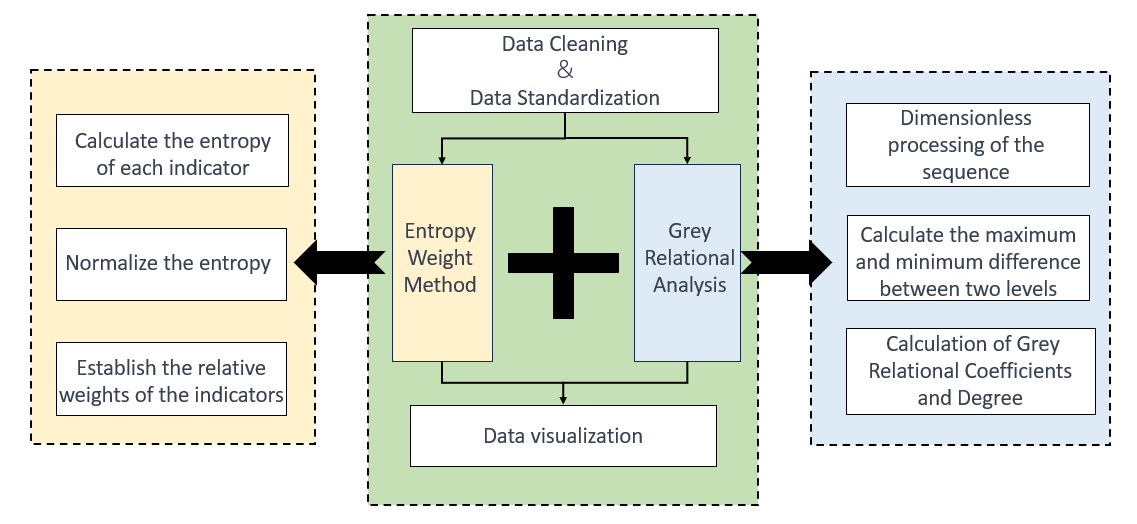}
	\caption{Solution Flowchart for Task 1}
\end{figure}

\subsection{EWM}
\noindent Entropy Weight Method(EWM) is an objective assignment method, which relies on the data itself to get the weights and can circumvent the influence of subjective factors.Therefore, given that there are n evaluation objects and m evaluation indicators, the weight of each indicator can be calculated through using EWM.We take the result indices obtained in Table 1 as the values of each dimension of the x vector in turn and each game as an object of evaluation. The concrete steps are as follows.

\begin{itemize}
\setlength{\parsep}{0ex} 
\setlength{\topsep}{2ex} 
\setlength{\itemsep}{1ex} 
\item \textbf{Step 1:} Construction of the raw data matrix.

$$
\tilde{X}=\left[ \begin{matrix}
	\tilde{x}_{11}&		\tilde{x}_{12}&		\cdots&		\tilde{x}_{1m}\\
	\tilde{x}_{21}&		\tilde{x}_{22}&		\cdots&		\tilde{x}_{2m}\\
	\vdots&		\vdots&		\ddots&		\vdots\\
	\tilde{x}_{n1}&		\tilde{x}_{n2}&		\cdots&		\tilde{x}_{nm}\\
\end{matrix} \right].,
$$
where $\tilde{x}_{ij}$ is the value of the $j_{th}$ indicator of the $i_{th}$ evaluation object, $i=1,2,\cdots,n;j=1,2,\cdots,m$; $n=62, m=10$.

\item \textbf{Step 2:} Standardlized processing of data.

\begin{equation}
x_{ij}=\frac{\tilde{x}_{ij}-\min \left\{ \tilde{x}_{1j},\tilde{x}_{2j},\cdots \tilde{x}_{nj} \right\}}{\max \left\{ \tilde{x}_{1j},\tilde{x}_{2j},\cdots ,\tilde{x}_{nj} \right\} -\min \left\{ \tilde{x}_{1j},\tilde{x}_{2j},\cdots ,\tilde{x}_{nj} \right\}}+\varepsilon,
\end{equation}
where $\varepsilon$ is an infinitesimal value.

\item \textbf{Step 3:} Calculation of the weight of indicators.

\begin{equation}
p_{ij}=\frac{x_{ij}}{\sum_{i=1}^{n}  x_{ij}},j=1,2,\cdots ,m.
\end{equation}

\item \textbf{Step 4:} Calculating the information entropy.

\begin{equation}
e_{j}=-\frac{1}{\ln n}\sum_{i=1}^{n} p_{ij}\ln \left( p_{ij} \right) ,j=1,2,\cdots ,m.
\end{equation}

\item \textbf{Step 5:} Getting the weight of the $j_{th}$ indicators.

\begin{equation}
\tilde{\omega} _j=\frac{1-e_j}{\sum_{i=1}^{n} \left( 1-e_i \right)}.
\end{equation}

\item\textbf{Step 6:} Introducing serve advantage factor \& Normalisation

\begin{equation}
\omega _j^{'}=\left\{ \begin{array}{l}
	\tilde{\omega _j}+\xi \ ,j=1\\
	\tilde{\omega _j}\ \ \ \ \ \ \ \ ,otherwise\\
\end{array} \right. 
\end{equation}
\noindent where$\ \xi \in \left[ 0.05,0.15 \right] ,\ \xi =0.1$.

\begin{equation}
\omega _j=\frac{\omega _j^{'}}{\underset{j=1}{\overset{m}{\varSigma}}\omega _j^{'}}.
\end{equation}
\end{itemize}
\begin{figure}[H]
	\centering
	\includegraphics[width=0.8\textwidth]{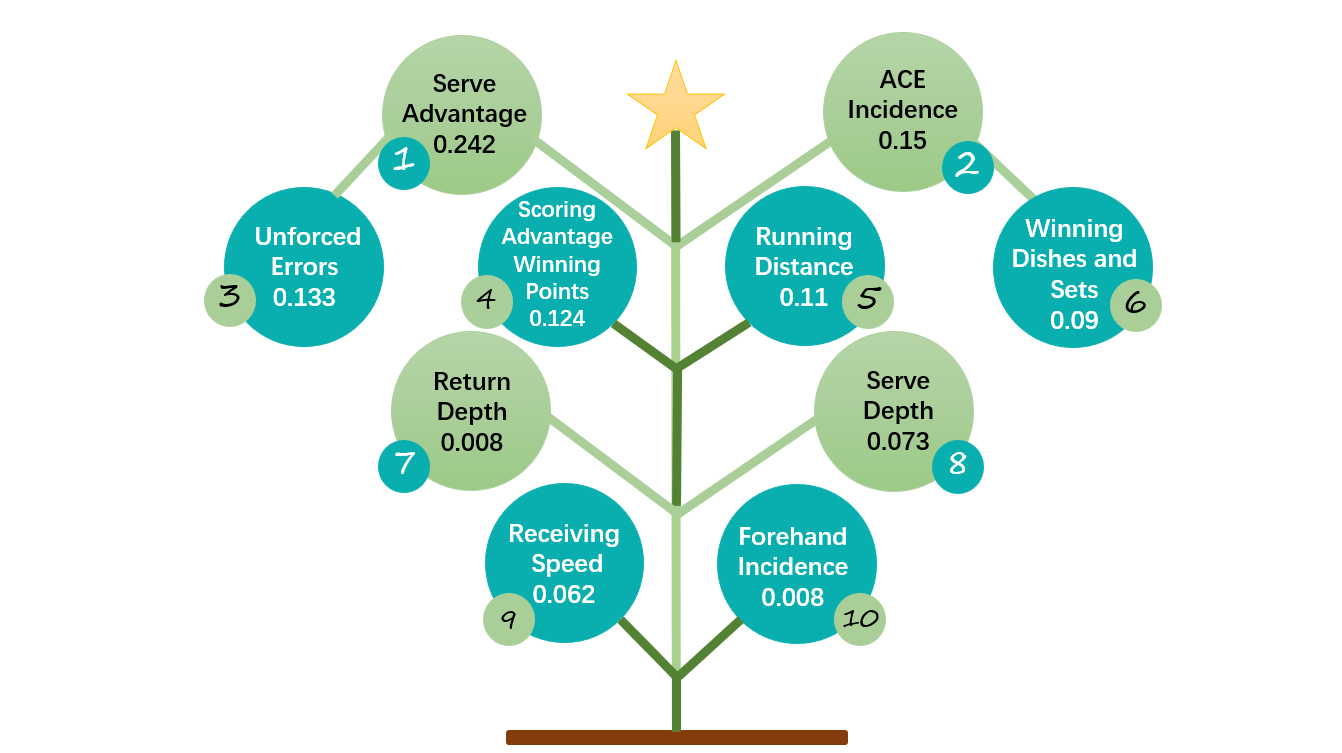}
	\caption{Calculated entropy weights}
\end{figure}


\subsection{GRA}
\noindent The entropy-grey relational analysis method is a technique that combines Entropy Weight Method(EWM) and Grey Relation Analysis(GRA) . It is used to introduce weights for different evaluation indicators, thereby providing a more objective assessment of each player's performance in the game.

\noindent Based on the grey system theory, the win-loss ratio of players in all matches is regarded as a grey system engineering. By introducing objective and reasonable entropy weights for each evaluation indicator on the basis of Grey Relation Analysis(GRA), the performance scores of each player at a specific time in the game are obtained. The detailed steps are as follows.

\begin{itemize}
\setlength{\parsep}{0ex} 
\setlength{\topsep}{2ex} 
\setlength{\itemsep}{1ex} 
\item \textbf{Step1:} Determine the analysis sequence.

\noindent Take the scoring margin of the game as the reference sequence, denoted as $X_0$. $X_i$ is the comparison sequence corresponding to the value of the previous $i_{th}$ evaluation metric.

\begin{equation}
	X_0=[x_0\left( 1 \right),x_0\left( 2 \right),\cdots,x_0\left( n \right)],
	X_i=[x_i\left( 1 \right),x_i\left( 2 \right),\cdots,x_i\left( n \right)],
\end{equation}
\noindent where $i=1,2,\cdots,m$.

\item \textbf{Step2:} Dimensionless processing of the sequence.

\begin{equation}
	x_i^{'}\left( k \right) =\frac{x_i\left( k \right)}{Mean_i},
\end{equation}
where $Mean_i$ denotes the mean of the $i_{th}$ column.

\item \textbf{Step3:} Calculate the maximum and minimum difference between two levels.

\begin{equation}
	\varDelta i\left( k \right) =|x^{'}_0\left( k \right) -x^{'}_i\left( k \right) |,\ i=1,2,\cdots ,m,
\end{equation}

\begin{equation}
	MA=\underset{i}{\max}\underset{k}{\max}\varDelta _i\left( k \right) , 
\end{equation}

\begin{equation}
	mi=\underset{i}{\min}\underset{k}{\min}\varDelta _i\left( k \right) .
\end{equation}

\item \textbf{Step4:} Calculation of Grey Relational Coefficients.

\begin{equation}
\gamma_{0i}(k) = \frac{mi + \xi MA}{\Delta_i(k) + \xi MA} , k=1,2,\cdots,m,
\end{equation}
where $\xi$ is the resolution factor, $\xi \in \left( 0,1 \right) $, generally $\xi$ takes the value of 0.5.

\item \textbf{Step5:} Calculation of Grey Relational Degree.

\begin{equation}
\Gamma_{0i} = \sum_{k=1}^{l} \omega_{i} \gamma_{0i}(k), \quad i = 1, 2, \ldots, m.
\end{equation}
\end{itemize}

\begin{table}[H]
\centering
\caption{Rank of weighted evaluation indicators by relevance}
\begin{tabular}{*{5}{c}}
\toprule
Indicators & Grey Relation & Weight & Weighted Relevance & Sequence \\
\midrule
	$x_{1}$&	0.924& 0.242& 0.115& 4 \\
	$x_{2}$&	0.926& 0.15& 0.224& 1 \\
	$x_{3}$&	0.918& 0.133& 0.122& 3 \\
	$x_{4}$&	0.917& 0.124& 0.083& 6 \\
	$x_{5}$&	0.917& 0.110& 0.0569& 7 \\
	$x_{6}$&	0.898& 0.09& 0.0071& 9 \\
	$x_{7}$&	0.858& 0.008& 0.063& 10 \\
	$x_{8}$&	0.904& 0.073& 0.099& 5 \\
	$x_{9}$&	0.936& 0.062& 0.140& 2 \\
	$x_{10}$&	0.933& 0.008& 0.0074& 8 \\
\bottomrule
\end{tabular}
\end{table}

\noindent In the actual score calculation every moment, the overall value of the specific time series to each time is analyzed, so we set a \textbf{sliding window} with a size of 10 to intercept the normalized value of the average data in the relevant time before this time as data input, denoted as \boldmath{$\overline{x_i}$}, and calculate the final performance score. The specific formula is as follows.
\begin{equation}
\boldmath{S=\underset{i=1}{\overset{m}{\varSigma}}\Gamma_{0i}\overline{x_i},i=1,2,\cdots,m.}
\end{equation}

\noindent Visualise the dynamic progression of the game based on the players' performance scores, with the following image.

\begin{figure}[H]
	\centering
	\subfigure[2023-wimbledon-1701]{
	\includegraphics[scale=0.5]{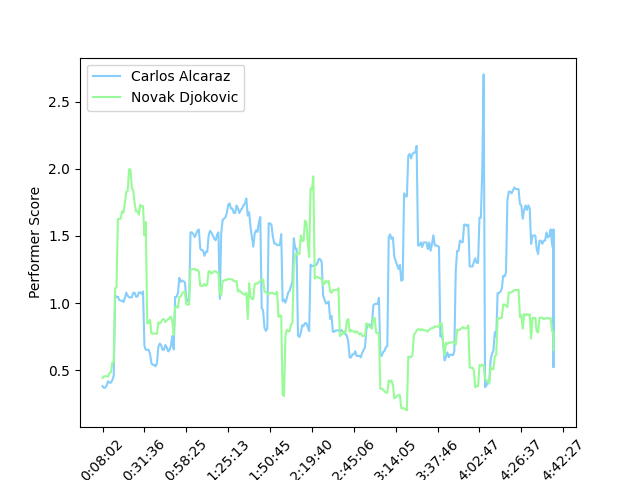}}
	\hspace{0.1in}
	\subfigure[2023-wimbledon-1401]{
	\includegraphics[scale=0.5]{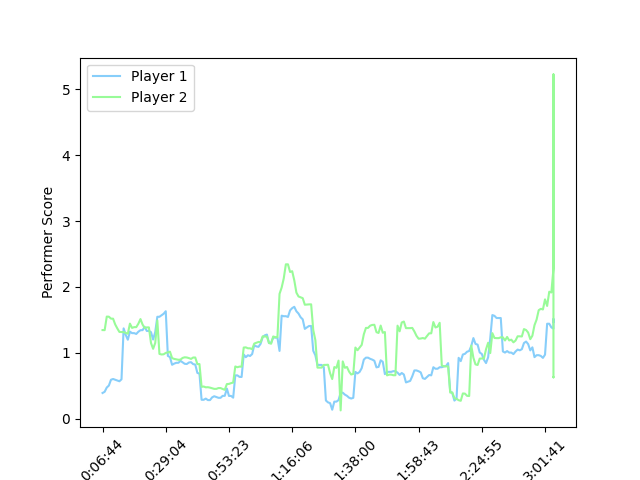}}
	\caption{Visualization of Match flow}
\end{figure}

\section{EMA-RT Validation Model}

\begin{figure}[H]
	\centering
	\includegraphics[width=1\textwidth]{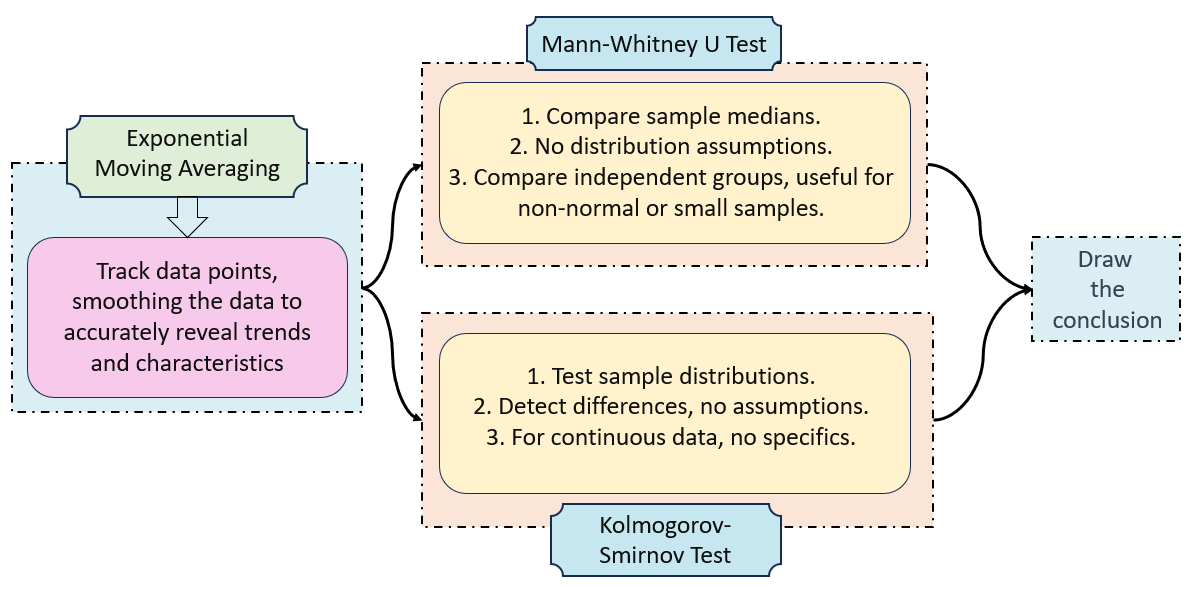}
	\caption{Solution Flowchart for Task 2}
\end{figure}

\subsection{EMA}

\noindent \textbf{Momentum} is a force that dictates the flow of the game. It cannot be directly visualised because it affects how a player performs at a given moment. However, we can feel momentum in the real-time movements and outcomes of the game. There are many seemingly twists and turns in the results of the first question, but the very small fluctuations cannot essentially reflect the momentum, that is, a changing trend. Therefore, we use Exponential moving averaging method to capture the momentum.Exponential moving averaging, also called weighted moving averaging, makes the corresponding image smoother and smoother by averaging the data by giving higher weights to the data in adjacent times. We utilize EWM-GRA model to calculate the performance scores of players, and use EMA to process the above time series. The following are the detailed steps.

\noindent Supposing that the data series of player performance scores in a match is $y_1, y_2, \cdots, y_n$,

\begin{equation}
EMA_t=\frac{y_t+\left( 1-\alpha \right) y_{t-1}+\left( 1-\alpha \right) ^{2}y_{t-2}+\cdots +\left( 1-\alpha \right) ^{t}y_0}{1+\left( 1-\alpha \right) +\left( 1-\alpha \right) ^{2}+\cdots +\left( 1-\alpha \right) ^{t}},
\end{equation}

\noindent where the sliding period $t=10s$, smoothing factor $0\le \alpha \le 1$, $\alpha=0.9$.

\noindent After exponential moving average and visualisation of the data, it can be seen that the performance scores of players in the process of the game will continuously increase or decrease in a certain period of time, which is due to the appearance of the \textbf{momentum }that causes continuous changes in the performance of the players. Thus, \textbf{we regard momentum as the trend} of a player's performance scores over a certain period of time. The concrete images are as follows.

\begin{figure}[H]
	\centering
	\subfigure[match1]{
	\includegraphics[scale=0.5]{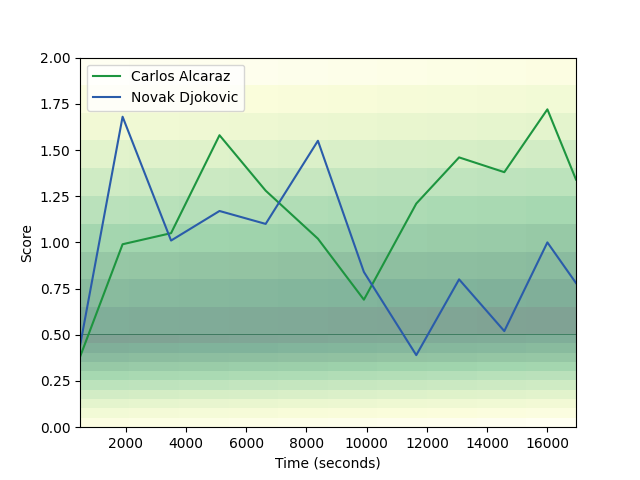}}
	\hspace{0.1in}
	\subfigure[match2]{
	\includegraphics[scale=0.5]{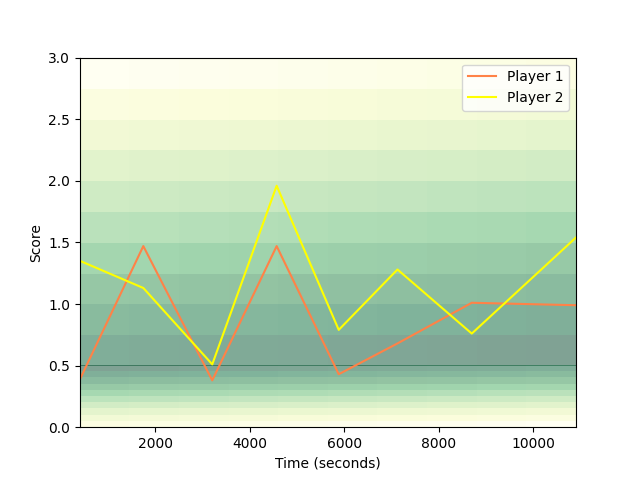}}
	\caption{EWA-processed match momentum}
\end{figure}

\subsection{Randomness Test}
\noindent Raw data is a collection of all players' game trends, and it represents situations where there is  \textbf{momentum} in a game. Given that some weights affect randomness, we presuppose a reasonable set of probabilities for each metric of the randomly generated dataset, and use a computer to generate 1,000 random datasets as new sample sets that match the original data to simulate the lack of \textbf{momentum} in the game.

\begin{figure}[H]
	\centering
	\includegraphics[width=0.7\textwidth]{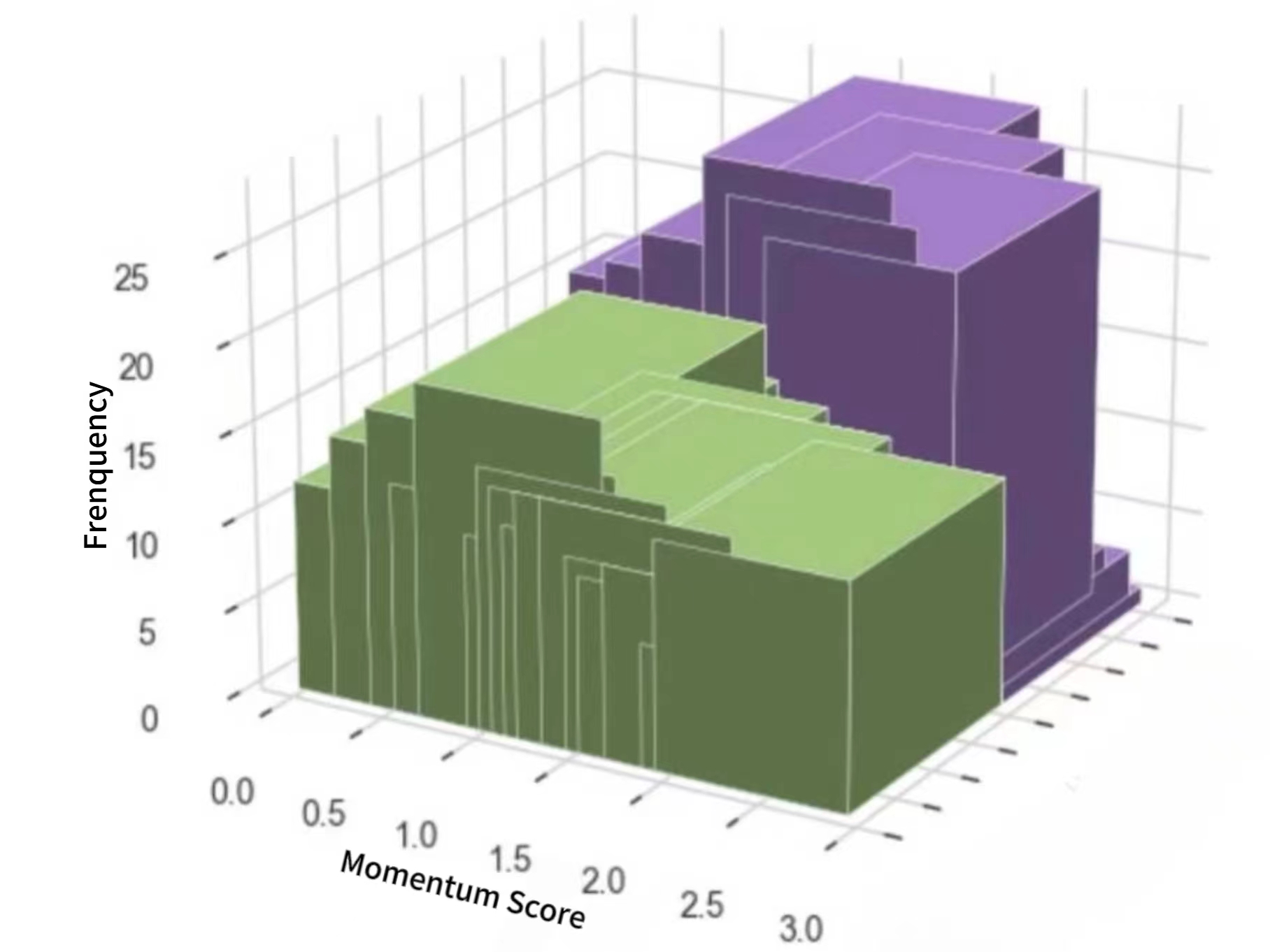}
	\caption{Distribution of Momentum Scores: Actual VS. Simulated}
\end{figure}

\noindent Confirming that the significance level \boldmath{$\alpha=0.05$ }and sample size \boldmath{$n_1=62, n_2=1000$}, the following hypothesis testing will be completed using Mann-Whitney U test\textsuperscript{\cite{2}} and Kolmogorov-Smirnov test\textsuperscript{\cite{3}} to analyse whether there is a correlation between the presence of momentum and the overall momentum of the race.Then set the hypotheses.

\begin{itemize}
\setlength{\parsep}{0ex} 
\setlength{\topsep}{2ex} 
\setlength{\itemsep}{1ex} 

\item Null hypothesis($H_0$): There is no significant difference between the overall performance of the group with and without "momentum" in the competition.
\item Alternative hypothesis($H_1$): The group with "momentum" in the competition performs significantly better than the group without "momentum" in the overall performance.
\end{itemize}

\subsubsection{Mann-Whitney U Test}

\begin{equation}
	\begin{aligned}
	U-statistic: U_1&=n_1n_2+\frac{n_1\left( n_1+1 \right)}{2}-R_1, \\
	U_2&=n_1n_2+\frac{n_2\left( n_2+1 \right)}{2}-R_2, \\ 
	\end{aligned}
\end{equation}

\begin{equation}
U=min\{U_1,U_2\},
\end{equation}

\noindent where \boldmath{$R_1,R_2$} respectively represent the rank sum of two sets of samples.

\noindent After checking the table to know the value of $U_\alpha$, we calculated \boldmath{$p-value=0.0043<0.05$}, then null hypothesis($H_0$) is rejected and alternative hypothesis($H_1$) is accepted, which states that players with Momentum in the game significantly outperform players without Momentum in overall performance.

\subsubsection{Kolmogorov-Smirnov Test}

\begin{equation}
Smirnov-statistic: D_{n_1,n_2}=\underset{x}{sup}|F_1\left( x \right) -F_2\left( x \right) |,
\end{equation}

\begin{equation}
\label{eq1}
D_{n_1,n_2}>c\left( \alpha \right) \sqrt{\frac{n_1+n_2}{n_1n_2}}.
\end{equation}

\noindent If inequality (\ref{eq1}) holds, null hypothesis($H_0$) will be rejected at level $\alpha$.After checking the table to know the value of $D_{n_1,n_2,\alpha}$ and calculating, we get \boldmath{$p-value=0.00128<0.05$}. Then it is shown that the above conclusion is also valid.
\noindent To sum up, we can see that after two tests of randomness, the P-values are both less than 0.05, and the H0 hypothesis can be rejected, that is, the emergence of momentum is not random with the trend of the game, but correlated.

\section{XGBoost-SHAP Prediction Model}

\begin{figure}[H]
	\centering
	\includegraphics[width=0.70\textwidth]{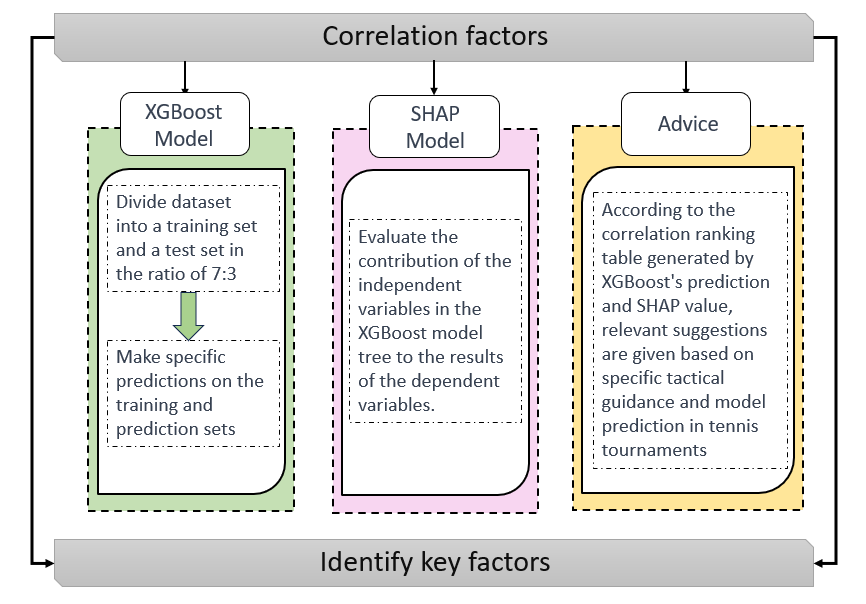}
	\caption{Solution Flowchart for Task 3}
\end{figure}

\subsection{XGBoost}
\noindent By observing the data, we found that the amount of data is large, and it is difficult to regression to a certain extent. After comprehensively considering the prediction and the correlation of multiple variables, we compared the accuracy of multiple algorithms of machine learning algorithms, and selected XGBoost regression analysis in machine learning as the prediction model.
\noindent XGBoost algorithm\textsuperscript{\cite{4,5}} is an integrated learning method based on the idea of Boosting. It helps in exploiting every bit of memory and hardware resources for tree boosting algorithms. The following is a summary graph of XGBoost based on the 2W+1H principle.

\vspace{0.5cm}
\noindent In order to construct the momentum prediction model, we need to consider the influence of various factors on the trend, not only in the evaluation metrics, but all potential influence factors related to the field. Combining the original given data, we combine all the variables of the original data (in the case of athlete p1, whose opponent is p2). These data were used as input data for XGBoost regression analysis.
\begin{figure}[H]
	\centering
	\includegraphics[width=0.9\textwidth]{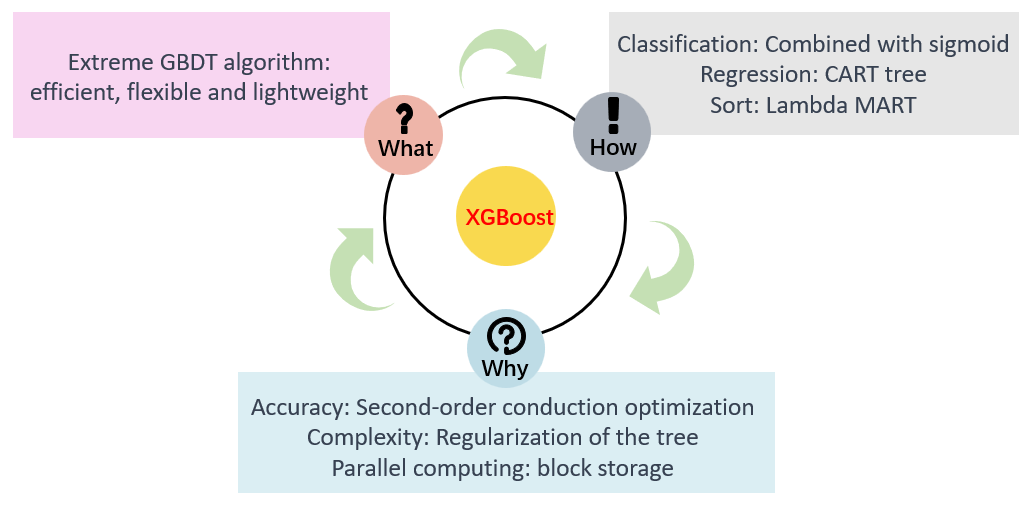}
	\caption{XGBoost summary graph based on 2W+1H}
\end{figure}
\noindent In order to construct the momentum prediction model, we need to consider the influence of various factors on the trend, not only in the evaluation metrics, but all potential influence factors related to the field. Combining the original given data, we combine all the factors of the original data (in the case of athlete p1, whose opponent is p2). These data were used as independent variable input data for XGBoost regression analysis.
\noindent We constitute a multi-dimensional dataset by combining independent variable input data with the performance scores of players in each match calculated using Model 1, and divide it into a training set and a test set in the ratio of 7:3. We trained XGBoost model using the training set and tested the model with the test set and Carlos Alcaraz and Novak Djokovic final data. After computational analysis, the accuracy of predicting fluctuations over multiple matches is $0.9990134$ and $0.992738$ for final(single match).The specific prediction results for training and prediction sets are as follows.

\begin{figure}[H]
	\centering
	\subfigure[multiple matches]{
	\includegraphics[scale=0.31]{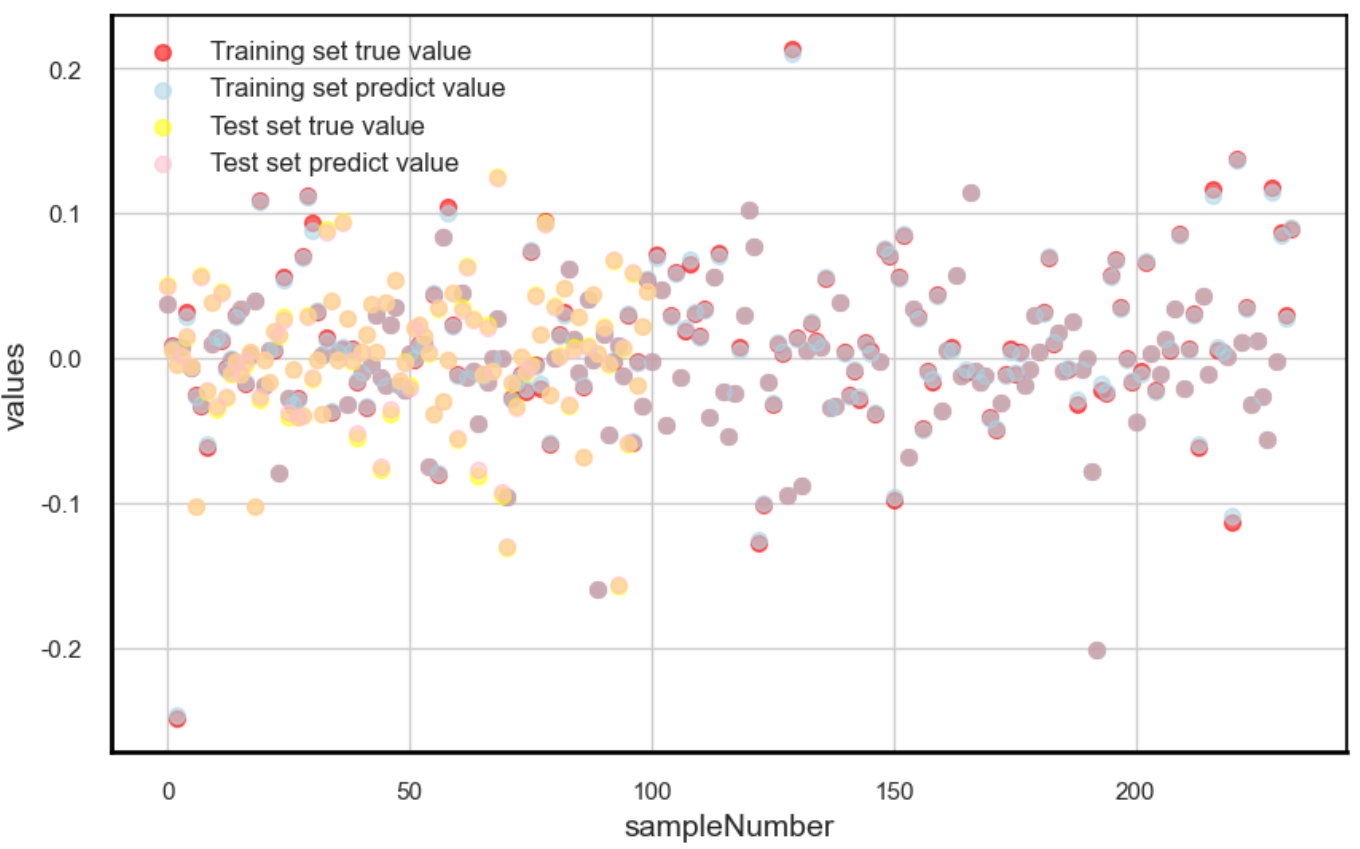}}
	\hspace{0.1in}
	\subfigure[final(single match)]{
	\includegraphics[scale=0.31]{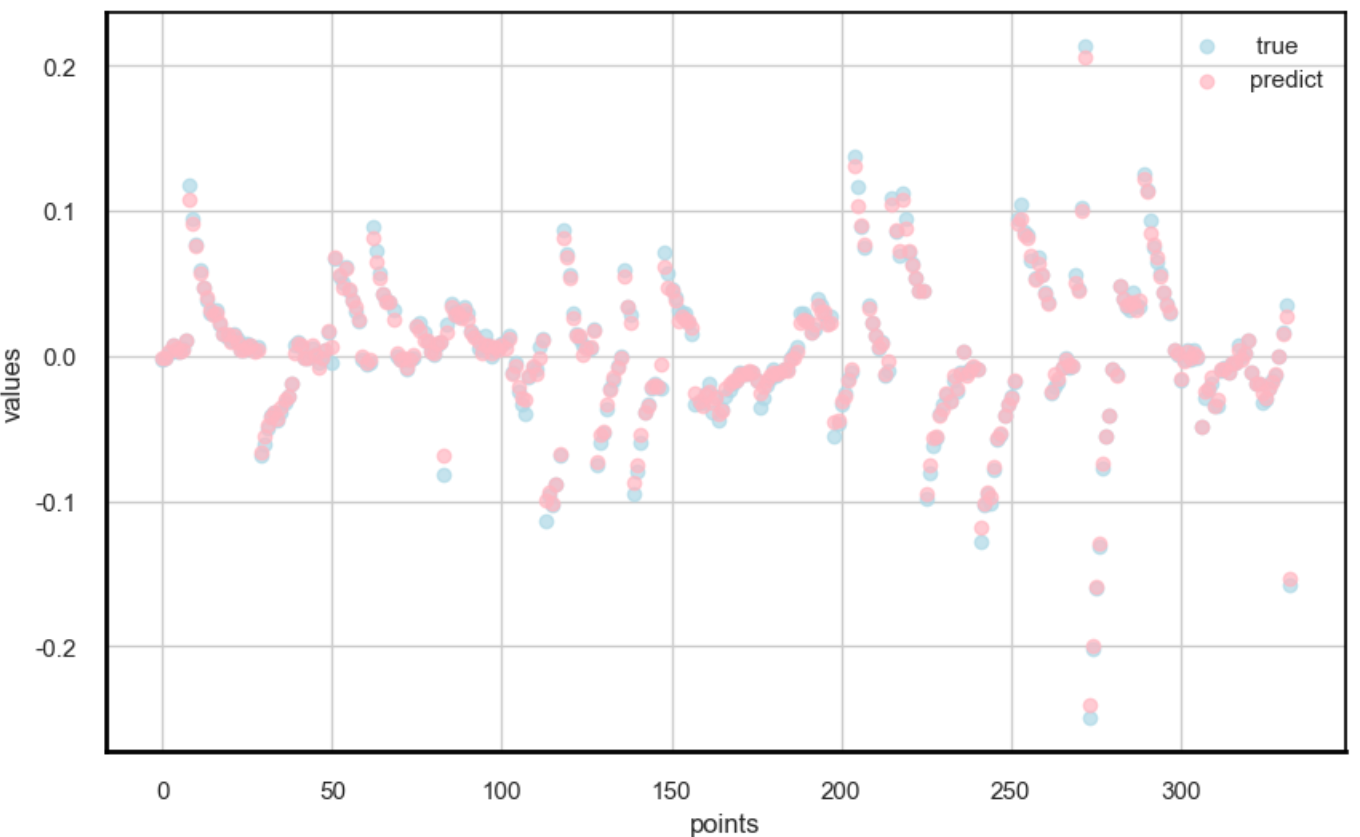}}
	\caption{True VS. Predicted}
\end{figure}

\subsection{SHAP}

\noindent For ensemble learning XGBoost model, the high accuracy is achieved by complex alogrithms, which creates a tension between accuracy and interpretability. In response, we apply a method from game theory, SHapley Additive exPlanations(SHAP)\textsuperscript{\cite{6,7}} to assign each factor an importance value for a particular prediction and evaluate the contribution of the independent variables to the outcome of the dependent variable in the tree of XGBoost model.

\begin{figure}[H]
	\centering
	\includegraphics[width=0.9\textwidth]{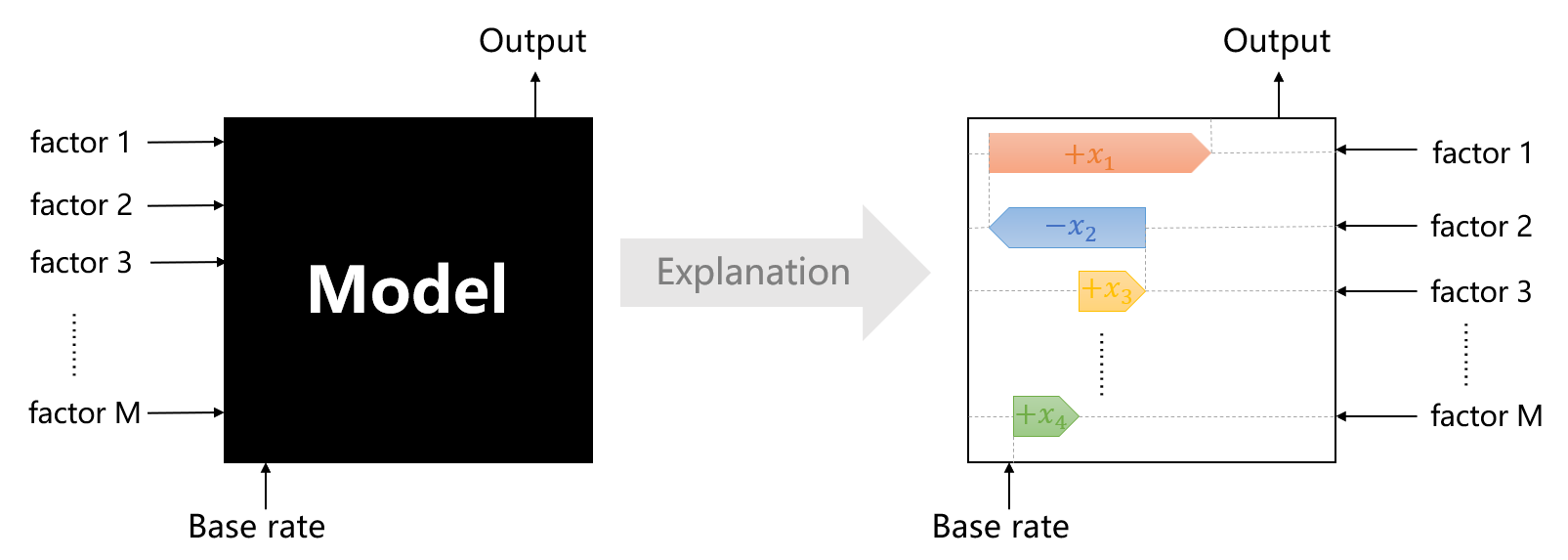}
	\caption{SHAP summary graph}
\end{figure}

\noindent After processing and calculation of SHAP, the final contribution of all factors is obtained. Top 15 factors and corresponding SHAP values are as follows(All factors' SHAP values are shown in Figure \ref{fig} in the appendix). These factors are the most relevant ones that can be obtained.
\textbf {As can be seen from the table below, distance ran during point by both sides seems most related with swings in the play.}
\begin{table}[H]
\centering
\caption{SHAP Values of top 15 facotrs}
\resizebox{\textwidth}{!}
{\begin{tabular}{c c}
\toprule[2pt]
\multicolumn{1}{m{5cm}}{\centering \textbf{Factors}}
&\multicolumn{1}{m{10cm}}{\centering \textbf{SHAP Values} }\\ 
\midrule
	$opponent\_distance\_run$&	0.1077 \\
	$distance\_run$&	0.1042 \\
	$opponent\_score$&	0.0898 \\
	$game\_no$&	0.0665 \\
	$point\_no$&	0.0626 \\
	$speed\_mph$&	0.0617 \\
	$score$&	0.0502 \\
	$winner\_shot\_type$&	0.0453 \\
	$serve\_width$&	0.0419 \\
	$game\_victor$&	0.0416 \\
	$games$&	0.0386 \\
	$break\_pt\_won$&	0.0336 \\
	$point\_victor$&	0.0234 \\
	$opponent\_points\_won$&	0.0220 \\
	$serve\_depth$&	0.0197 \\
\bottomrule
\end{tabular}}
\end{table}

\noindent We also analysed the possible interactions between various factors. The followings are a graphical representation of the interactions between some of representative factors.

\begin{figure}[H]
	\centering
	\subfigure[$F_{1}$ and $F_{2}$]{
	\includegraphics[scale=0.42]{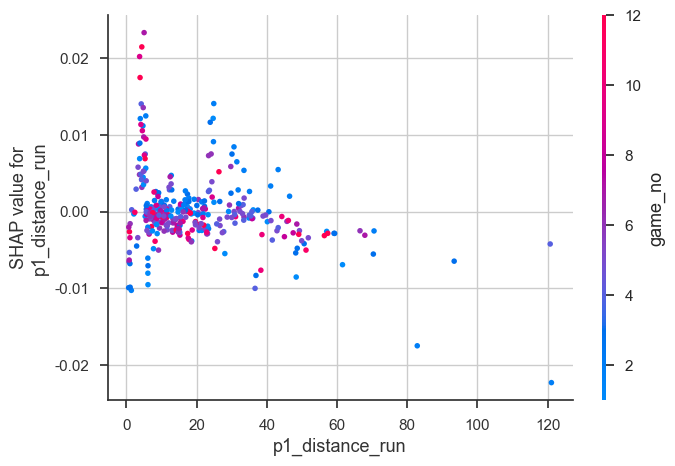}}
	\hspace{0.1in}
	\subfigure[$F_{3}$ and $F_{4}$]{
	\includegraphics[scale=0.42]{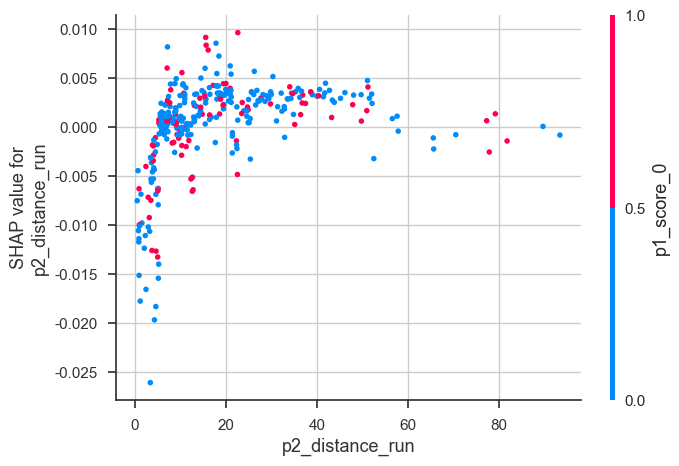}}
	\subfigure[$F_{6}$ and $F_{5}$]{
	\includegraphics[scale=0.42]{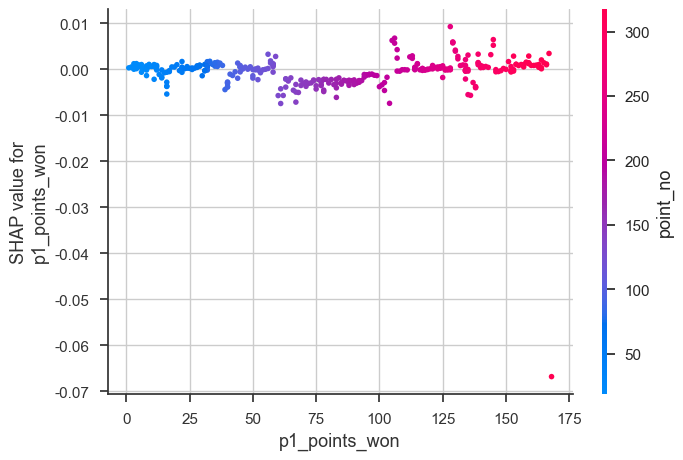}}
	\hspace{0.1in}
	\subfigure[$F_{7}$ and $F_{5}$]{
	\includegraphics[scale=0.41]{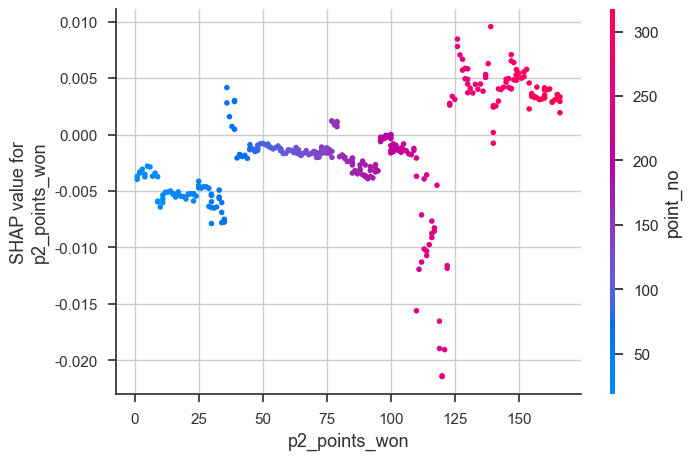}}
	\subfigure[$F_{5}$ and $F_{6}$]{
	\includegraphics[scale=0.41]{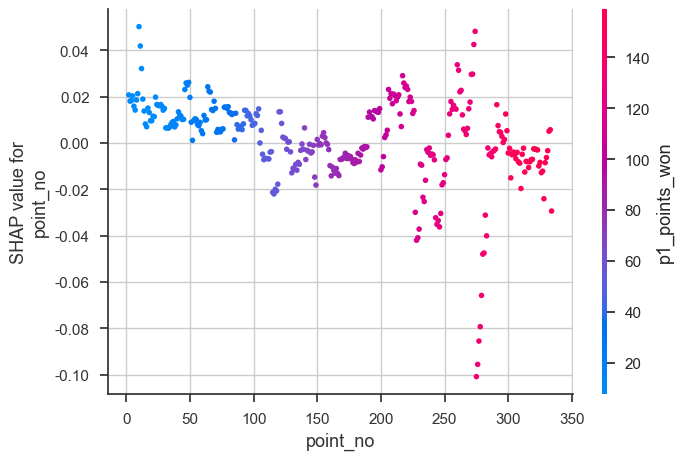}}
	\hspace{0.1in}
	\subfigure[$F_{2}$ and $F_{8}$]{
	\includegraphics[scale=0.42]{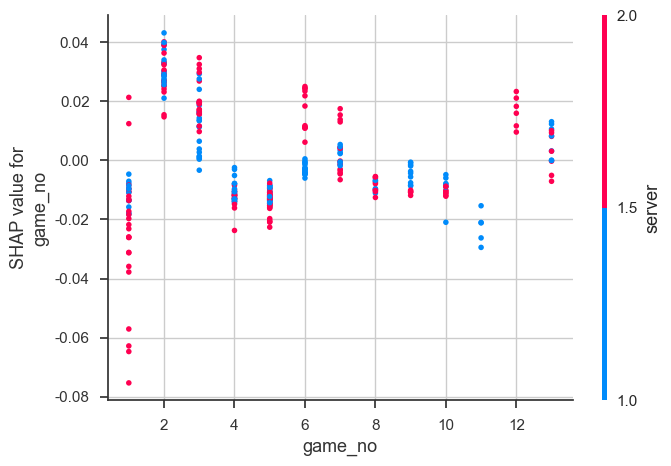}}
	\caption{Interaction diagrams between factors}
\end{figure}

\noindent The SHAP values are similar to regression coefficients, which can be positive or negative, large or small. The stronger the correlation between two factors, the more dispersed SHAP values are. The colour bar on the right side of the graph corresponds to SHAP values, with red representing a positive influence and blue representing a negative influence. According to the above figure, it can be seen that the interaction between $F_{1}$ and $F_{2}$, $F_{7}$ and $F_{4}$, $F_{2}$ and Server is not strong; at small values of $F_{6}$, it affects $F_{5}$ negatively, and when $F_{6}$ reaches a certain value, it affects $F_{5}$ positively, and the similar relationship of this kind is found between $F_{7}$ and $F_{5}$, and $F_{5}$ and $F_{6}$.Please see figure \ref{figa} in the appendix for the impact scores and correlation of the model parameters under different SHAP values in the prediction model training.

\section{ Generalisability Test}
\vspace{-0.5cm}
\begin{figure}[H]
	\centering
	\includegraphics[width=0.7\textwidth]{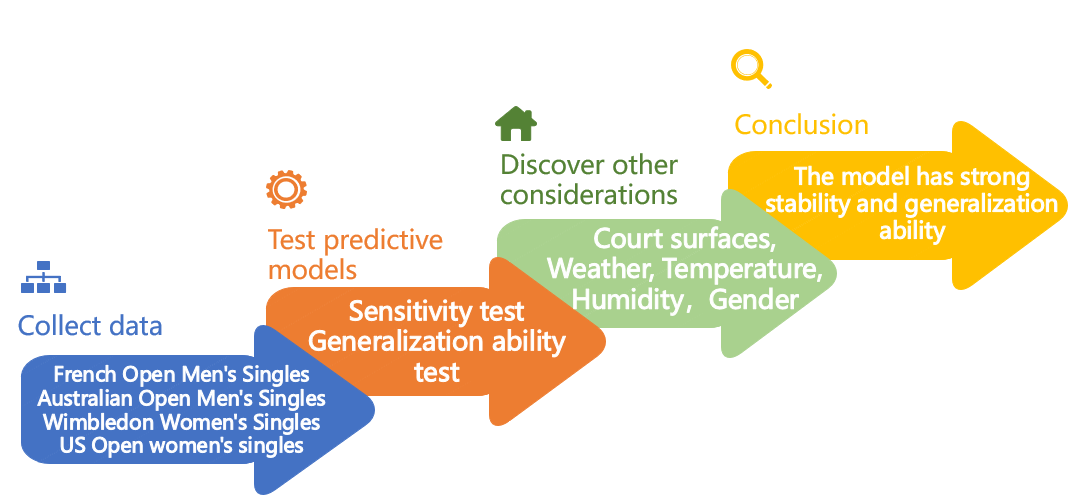}
	\caption{Solution Flowchart for Task 4}
\end{figure}

\subsection{Generalisability Test}
\noindent The prediction results of swings in the match of original data  have been presented in Task 3. Considering different court surfaces, French Open matches are on clay, Wimbledon is on grass, and US Open and Australian Open are on hard courts, whereas the court surface of Wimbledon in the raw data is always grass. Therefore, in order to conduct Generalisability Test\textsuperscript{\cite{8}} of the model, we collected data from men's matches at French Open and Australian Open in 2023 to analyse the effect of court surfaces on the model's prediction results. Meanwhile, women's matches at Wimbledon and US Open were chosen to test the effect of gender on model's prediction results. Then we fed these match data into the above prediction model and the following are the results of model's predictions.

\begin{figure}[H]
	\centering
	\subfigure[French Open men's singles]{
	\includegraphics[scale=0.21]{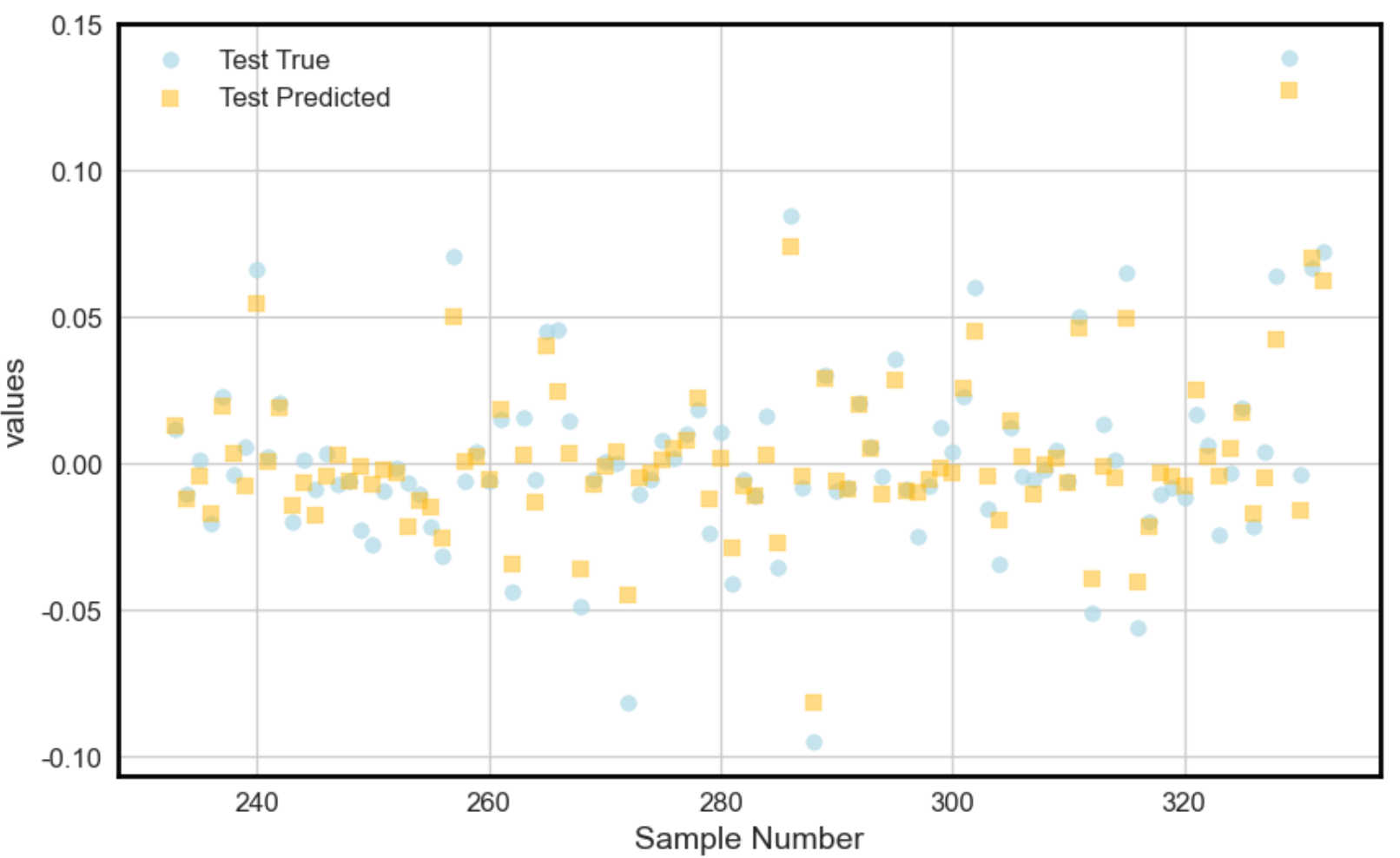}}
	\hspace{0.1in}
	\subfigure[Australian Open men's singles]{
	\includegraphics[scale=0.21]{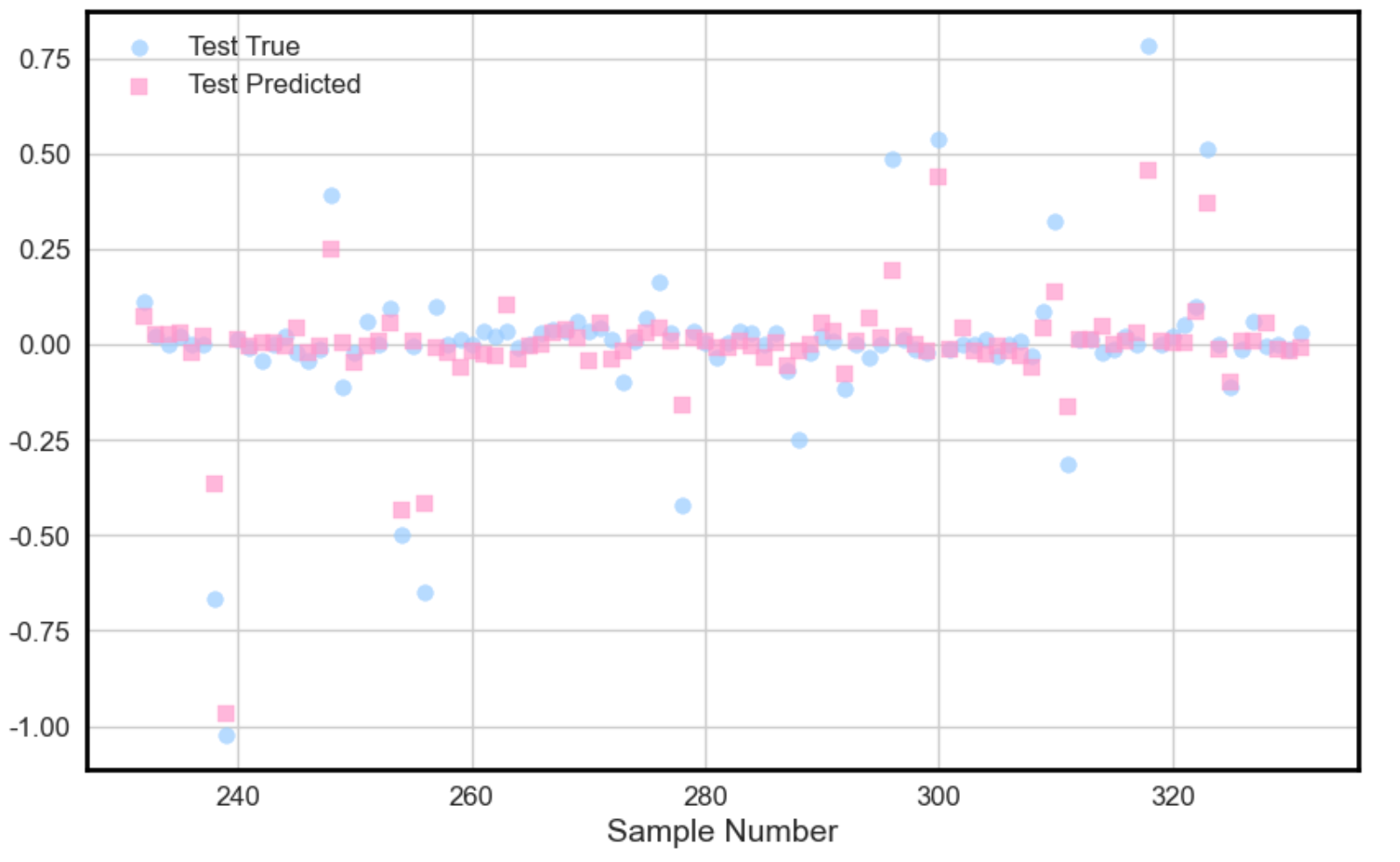}}
	
	\subfigure[Wimbledon women's singles ]{
	\includegraphics[scale=0.21]{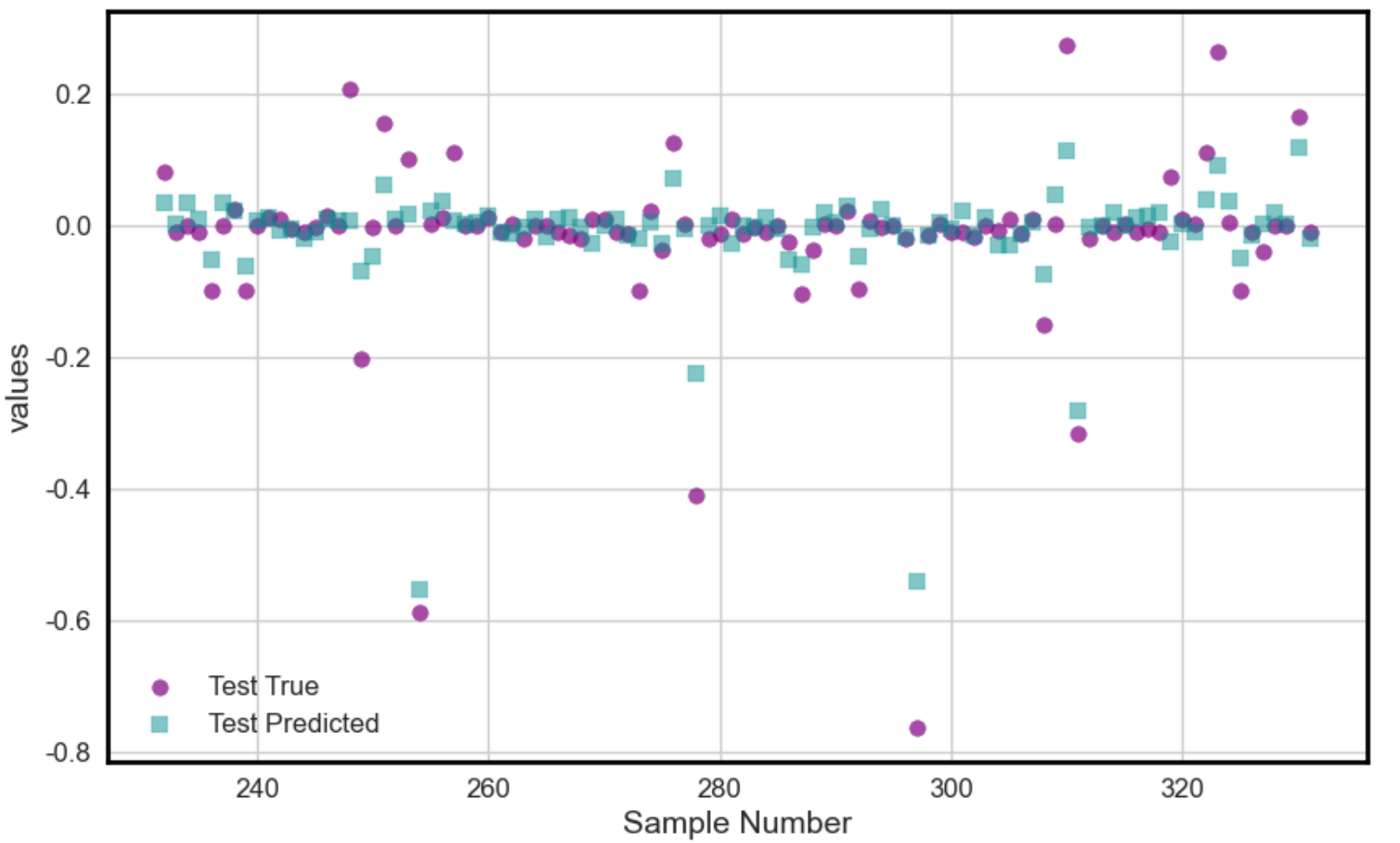}}
	\hspace{0.1in}
	\subfigure[US Open women's singles]{
	\includegraphics[scale=0.21]{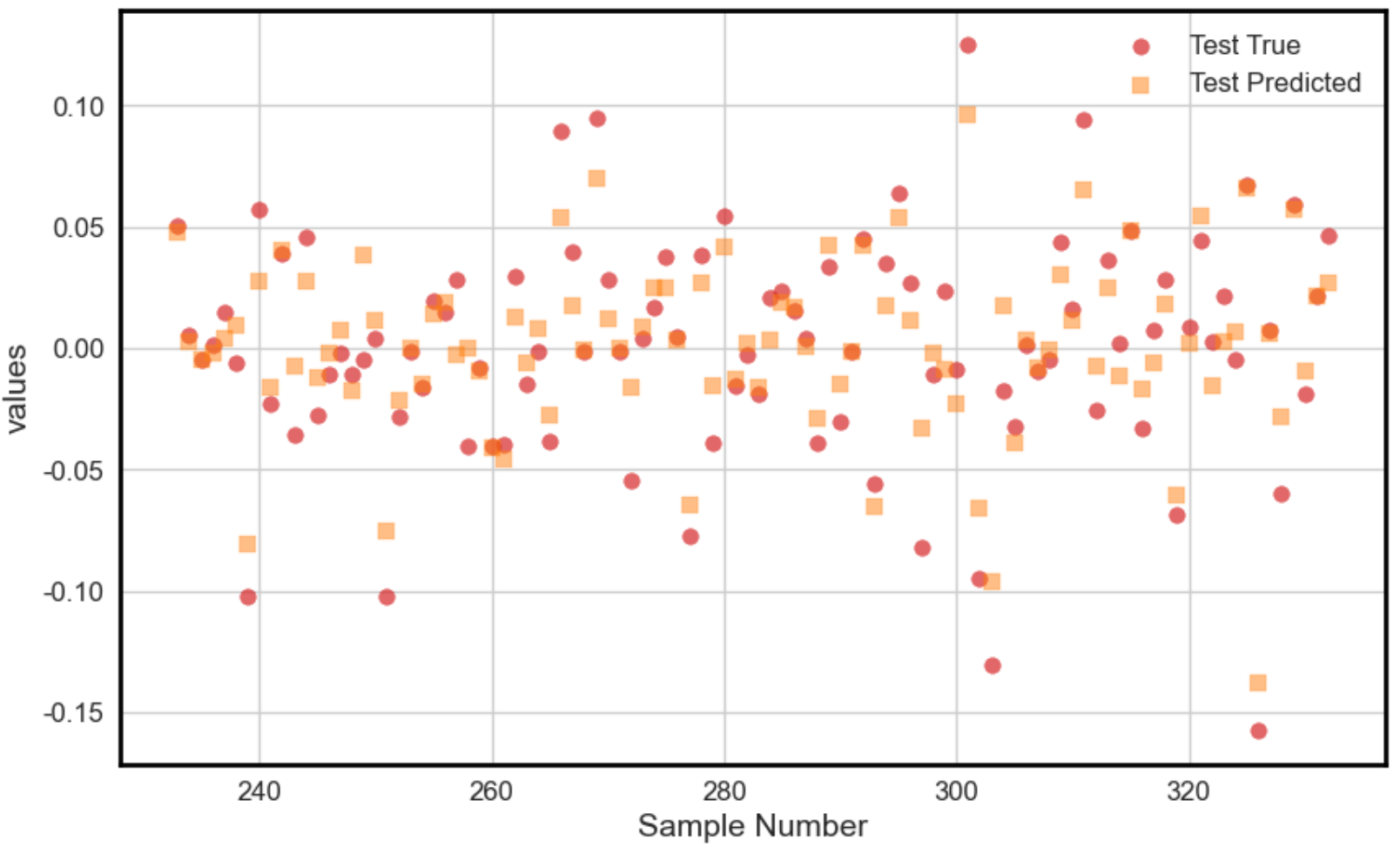}}
	\label{True VS. Predicted}
\end{figure}

\noindent 
As shown in the above figure, the closer predicted value to the true value is , the more concentrated the scatterplot is and the higher the accuracy is. Compared to the model's prediction accuracy $0.9990$ for Wimbledon men's singles, the model's prediction accuracy for French Open men's, Australian Open men's, Wimbledon women's, and U.S. Open women's matches have slightly decreased by various degrees, respectively of $0.9112, 0.9467, 0.9651, 0.8620$.

\subsection{Conclusion}

\begin{itemize}
\item \noindent The magnitude of generalisation ability of prediction model is examined by testing whether the model performs well for a new test set using match data with different factors such as court surfaces and gender. The final accuracy calculated shows that our model is robust and has a high level of prediction ability for tennis matches in other scenarios.

\item \noindent Admittedly, given that the model sometimes can be less accurate in predicting swings in other match, we should consider adding factors such as court surfaces, Off-site factors at match venues such as weather, temperature and humidity where the match is held, gender and other factors to future models to refine the parameters of the model.
\end{itemize}
\vspace{0.6cm}

\noindent{\setlength{\parindent}{2em}
 \noindent \setlength{\parindent}{2em}After carefully selecting key factors and analyzing 
 \noindent \setlength{\parindent}{2em}their correlation with momentum using SHAP, we have 
 \noindent \setlength{\parindent}{2em}concluded that \textbf{momentum} is a pivotal \noindent aspect of a tennis player's performance. 
 \noindent \setlength{\parindent}{2em}Consequently, before encountering a new opponent, 
 \noindent \setlength{\parindent}{2em}we offer the following comprehensive recommendations to the player:  }
\vspace{0.6cm}
 
\noindent \textbf{Strategic Movement and Shot Selection:}

\begin{itemize}  
       \item When behind in points, the player should aim to elongate the opponent's movement by executing lengthy, cross-court shots, effectively engaging them and exposing weaknesses.  
       \item Upon establishing a scoring edge, the player should minimize excessive movement to preserve energy and consolidate the advantage.  
\end{itemize}  
  
\noindent \textbf{Mental Fortitude:}

    \begin{itemize}  
        \item The player must maintain a composed mindset, with a particular emphasis on heightened focus at the commencement and conclusion of the match.  
        \item When in the lead, it's crucial to remain aware of the opponent's score to prevent complacency; conversely, when trailing, it's essential to limit distractions from the opponent's score and retain confidence in one's strategy.  
    \end{itemize}  
  
\noindent \textbf{Physical Stamina and Rhythm Management:}
    \begin{itemize}  
        \item It's imperative to allocate physical energy prudently, ensuring consistent return and serve velocities throughout the entirety of the match, thereby effectively managing the game's rhythm.  
    \end{itemize}

\noindent \textbf{Enhancing Backhand Execution:}
    \begin{itemize}  
        \item The backhand shot is instrumental in gaining and maintaining an upper hand. Players should prioritize refining the quality of their backhand strokes to augment the consistency and aggression of their scoring.  
    \end{itemize}  
  
\noindent \textbf{Tactical Versatility and Error Reduction:} 
    \begin{itemize}  
        \item When trailing, it's prudent to experiment with varying shot depths, increase volleys and rallies near the net, and simultaneously bolster serve stability to minimize unforced errors, such as double faults.  
    \end{itemize}  
  
\noindent \textbf{Data Analysis and Strategic Planning:}
    \begin{itemize}  
        \item Gather historical data on the opponent, encompassing insights on backhand techniques (one-handed or two-handed), shot preferences, the ratio of long to short shots, movement patterns, preferred serve and return landing zones, and the proficiency of their forehand and backhand strokes.  
        \item Utilize this amassed data, in conjunction with our model's predictions of both players' performances, to conduct a thorough examination of the factors influencing momentum and craft a more refined match strategy.  
    \end{itemize}  
  
\noindent \textbf{Real-time Adjustments and Optimization:}
    \begin{itemize}  
        \item Continuously collect and scrutinize live data during the match to adapt variable indicators based on evolving circumstances.  
        \item Obtain optimal strategies through rigorous simulation analysis and nimbly adjust technical and tactical approaches to either sustain or alter the momentum of the contest.  
    \end{itemize}  
\vspace{-0.5cm}
\section{Stability Analysis}

\begin{itemize}
\item \noindent For the serve advantage factor $\xi$ in the EWA-GRA Evaluation Model, we regard it as a constant $\xi=0.1$ when introduced. However, in reality, due to the different levels of serve skills of various players, the value of $\xi$ may fluctuate within a certain range. Therefore, we will examine the impact of fluctuations in the serve advantage factor $\xi$ on final outputs of the model. We calculated one player's performance scores by using the model at each given moment corresponding to different values of $\xi$. The specific curves are as follows.

\begin{figure}[H]
	\centering
	\includegraphics[width=0.9\textwidth]{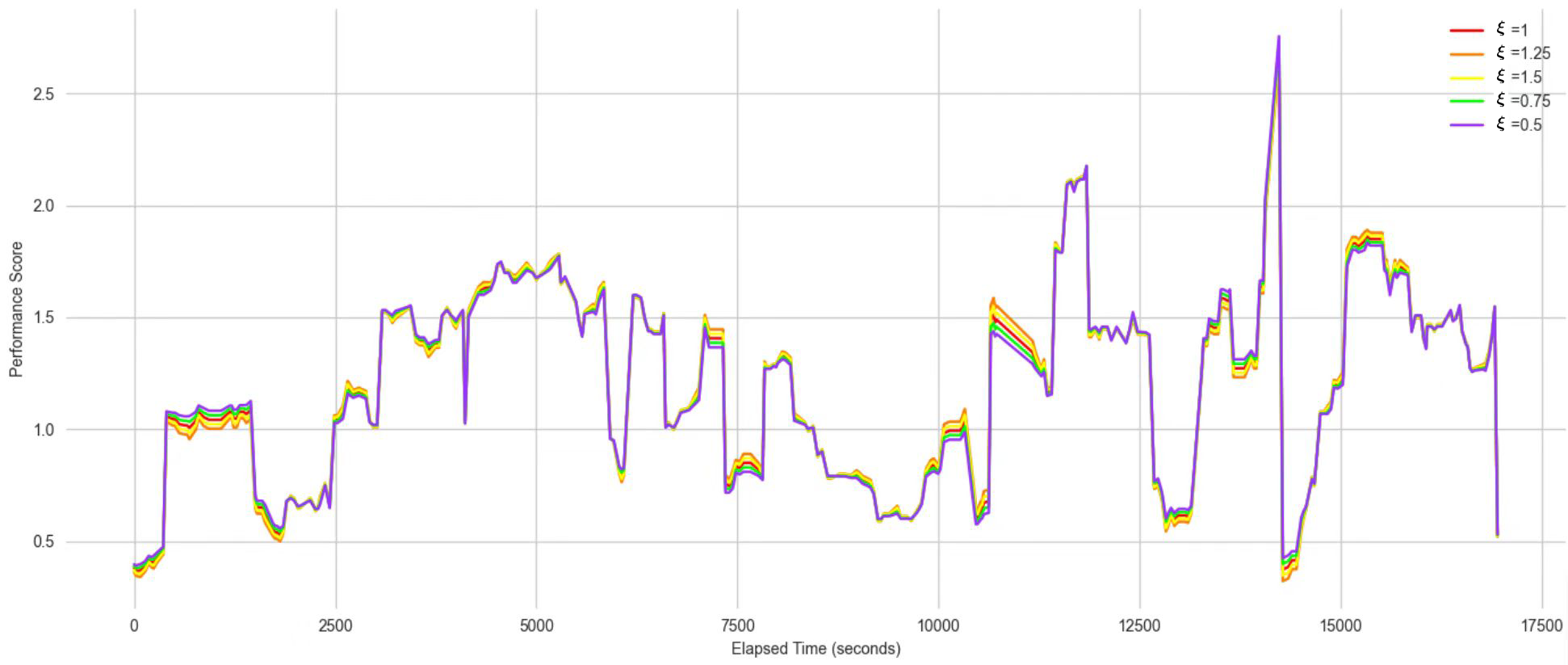}
	\caption{Performance scores curve corresponding to $\xi$ at each moment}
\end{figure}

\noindent After calculated, the change in the value of A causes the extreme variance of the model output to be $0.02579$ and the variance of the difference in the model output to be $0.00058$. Based on the calculated results and the trend of curves in the graph above, it can be shown that even if there is some slight error in the estimation of serve advantage factor $\xi$, the output of our evaluation model is still robust, (i.e. it is almost an optimal quantization of the player's performance score at a given moment).
\vspace{0.2cm}
\item \noindent For XGBoost-SHAP Prediction model, we have applied SHAP algorithm to analyze the impact and contribution of each factor to the outcome. The specific results are shown in figure \ref{figa} in the Appendix. The effect of changing the parameters on the accuracy of the model, the larger the absolute value of the SHAP value(i.e. the larger the contribution value), the more sensitive the model is to changes in the system parameters, the less stable the model is. Hence, the trend of the factor's influence on the model can be judged by the color of the corresponding points of each value. As shown by the calculation of SHAP algorithm and the results of Task 4, our prediction model has low sensitivity\textsuperscript{\cite{9}} and high generalization.

\end{itemize}
\section {Model Evaluation and Promotion}

\subsection{Strengths}
\begin{itemize}
\item
\noindent\textbf{EWA-GRA Evaluation Model:} Overcomes the limitations of the traditional mathematical evaluation model method, makes full use of all the information of each index, improves the accuracy of the evaluation conclusion, and is equally applicable to variations in sample size and sample irregularity.  
\item\noindent\textbf{EMA-RT Validation Model:} The Mann-Whitney U Test and Kolmogorov-Smirnov test were used to test the data after exponential averaging, and the accuracy of the model was verified according to the results.  
\item\noindent\textbf{XGBoost-SHAP Prediction Model:} Compared with other machine learning models, it is more robust and stable, which is more conducive to improving prediction accuracy.  
\item\noindent\textbf{Generalisability Test:} Testing on multiple sets of data makes the model more convincing.  
\end{itemize}  

\subsection{Weakness}
\begin{itemize}
\item\noindent\textbf{High Data Quality Requirements:} The accuracy and completeness of the data have a great impact on the results. If there are errors or missing data, the final decision results will also be affected, resulting in inaccurate decision-making.  
\item\noindent\textbf{Strict Model Assumptions:} The interpretation of nonlinear or complex models may not be accurate enough. There may be psychological factors or other factors affecting the results.  
\end{itemize}

\subsection{Promotion}
\noindent For other sports, by analyzing the SHAP correlation diagram and combining the key correlation factors of the model, such as \textbf{table tennis and other tennis-like sports}, the evaluation model and prediction model should be adjusted and trained at the same time under the changes of serve rules and venues. The data collection and adjustment of the weighting coefficients should be carried out for the factors with higher correlation in the model, such as increasing the opponent's running distance, increasing the depth of return and the score advantage. For the model itself, the changes in data training will \textbf{not affect its robustness and stability}, and from the data, the core indicators should be promoted and generalized, and applied to each game to formulate appropriate tactical arrangements and suggestions for the players, as well as to analyze the players' history of games before the start of the game to set the direction for training. \textbf{In many more areas, the description of momentum helps us to better understand the changes in sports games and capture the power of momentum.}

\clearpage   

\clearpage

\appendixpage
\setcounter{figure}{0}
\setcounter{table}{0}
\setcounter{section}{0}

\begin{figure}[H]
	\centering
	\includegraphics[width=0.9\textwidth]{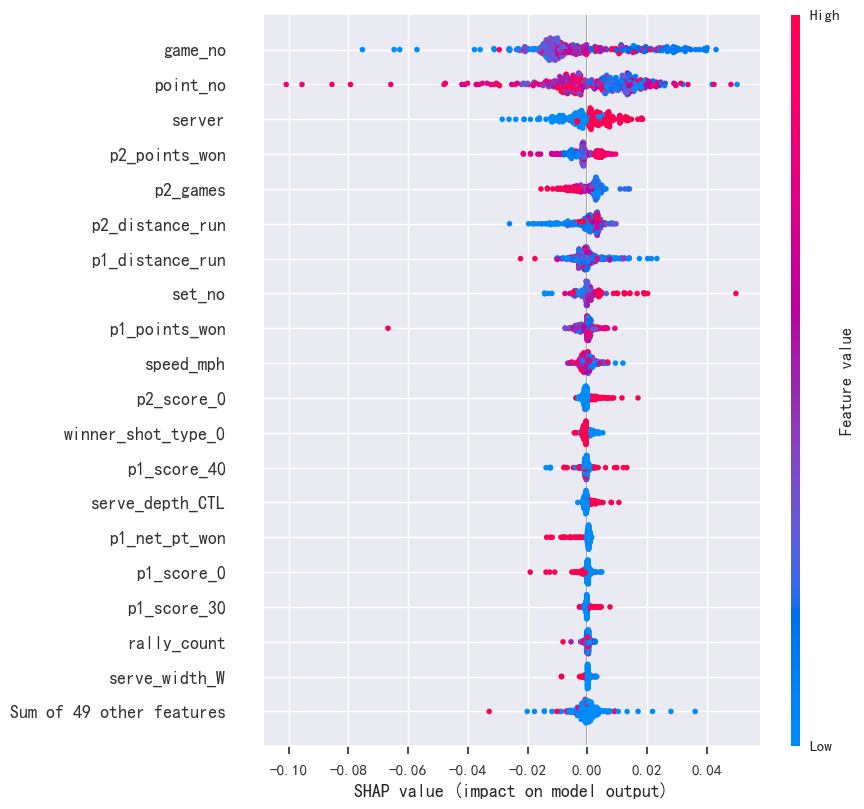}
	\caption{SHAP values \& correlation chart of factors with predicted outcomes}
	\label{figa}
\end{figure}

\begin{figure}[H]
	\centering
	\includegraphics[width=0.8\textwidth]{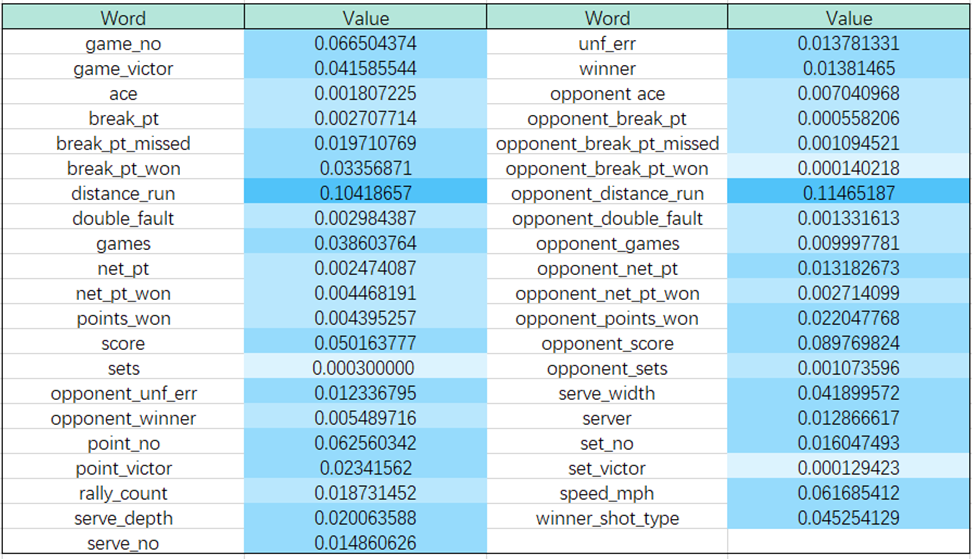}
	\caption{All Factors' SHAP Values}
	\label{fig}
\end{figure}


\begin{thebibliography}{1}

\bibitem{4}
Tianqi Chen and Carlos Guestrin.
\newblock Xgboost: A scalable tree boosting system.
\newblock In {\em Proceedings of the 22nd ACM SIGKDD International Conference on Knowledge Discovery and Data Mining}, KDD '16', page 785–794, New York, NY, USA, 2016. Association for Computing Machinery.

\bibitem{1}
Alistair Higham.
\newblock Momentum - the hidden force in tennis.
\newblock 2000.

\bibitem{3}
Joseph~L. Hodges and Erich~Leo Lehmann.
\newblock The efficiency of some nonparametric competitors of the t-test.
\newblock {\em Annals of Mathematical Statistics}, 27:324--335, 1956.

\bibitem{8}
Zhehan Jiang, Mark~R. Raymond, Christine Distefano, Dexin Shi, Ren Liu, and Junhua Sun.
\newblock A monte carlo study of confidence interval methods for generalizability coefficient.
\newblock {\em Educational and Psychological Measurement}, 82:705 -- 718, 2021.

\bibitem{6}
Scott~M. Lundberg and Su-In Lee.
\newblock A unified approach to interpreting model predictions.
\newblock In {\em Neural Information Processing Systems}, 2017.

\bibitem{7}
Mendbayar Otgonbayar, Badarifu, T.~Ranatunga, Takeo Onishi, and Ken Hiramatsu.
\newblock Cellular automata modelling approach for urban growth.
\newblock {\em Robotics and Autonomous Systems}, 6:93--104, 2018.

\bibitem{2}
K.~Vengatesan, S.~B. Mahajan, P.~Sanjeevikumar, R.~Mangrule, V.~Kala, and Pragadeeswaran.
\newblock {\em Performance Analysis of Gene Expression Data Using Mann--Whitney U Test}, pages 701--709.
\newblock Springer Singapore, Singapore, 2018.

\bibitem{9}
Pengfei Wei, Jingwen Song, Zhenzhou Lu, and Zhu feng Yue.
\newblock Time-dependent reliability sensitivity analysis of motion mechanisms.
\newblock {\em Reliab. Eng. Syst. Saf.}, 149:107--120, 2016.

\bibitem{5}
Ping Zhang, Yiqiao Jia, and Youlin Shang.
\newblock Research and application of xgboost in imbalanced data.
\newblock {\em International Journal of Distributed Sensor Networks}, 18, 2022.

\end{thebibliography}
\end{document}